\documentclass{article}

\usepackage{kafkanotes}

\usepackage[acronym, nomain, nonumberlist]{glossaries}
\makenoidxglossaries

\usepackage{natbib}

\usepackage{amsmath}
\usepackage{amssymb}
\usepackage{mathtools}
\usepackage{amsthm}
\usepackage{enumitem}
\usepackage{dsfont}

\usepackage{tikz}
\usepackage{pgfplots}
\pgfplotsset{compat=newest}
\usepackage{subcaption}
\usetikzlibrary{shapes.geometric}

\usepackage{tikz}
\usetikzlibrary{backgrounds}
\usetikzlibrary{arrows,shapes}
\usetikzlibrary{tikzmark}
\usetikzlibrary{calc}

\usepackage{amsmath}
\usepackage{amsthm}
\usepackage{amssymb}
\usepackage{mathtools, nccmath}
\usepackage{wrapfig}
\usepackage{comment}

\usepackage{blindtext}

\usepackage{graphicx}

\usepackage{xspace}

\usepackage{array}
\usepackage{ragged2e}
\newcolumntype{P}[1]{>{\RaggedRight\hspace{0pt}}p{#1}}
\newcolumntype{X}[1]{>{\RaggedRight\hspace*{0pt}}p{#1}}

\usepackage{tcolorbox}

\usepackage{tikz}
\usetikzlibrary{arrows,shapes,positioning,shadows,trees,mindmap}
\usepackage[edges]{forest}
\usetikzlibrary{arrows.meta}
\colorlet{linecol}{black!75}
\usepackage{xkcdcolors} 

\usepackage{tikz}
\usetikzlibrary{backgrounds}
\usetikzlibrary{arrows,shapes}
\usetikzlibrary{tikzmark}
\usetikzlibrary{calc}
\newcommand{\highlight}[2]{\colorbox{#1!17}{$\displaystyle #2$}}

\renewcommand{\highlight}[2]{\colorbox{#1!17}{#2}}

\newacronym{rl}{RL}{Reinforcement Learning}
\newacronym{drl}{DRL}{Deep Reinforcement Learning}
\newacronym{mdp}{MDP}{Markov Decision Process}
\newacronym{rpe}{RPE}{Reward Prediction Error}
\newacronym{pomdp}{POMDP}{Partially Observable Markov Decision Process}
\newacronym{td}{TD}{Temporal Difference}
\newacronym{per}{PER}{Prioritized Experience Replay}
\newacronym{deup}{DEUP}{Direct Epistemic Uncertainty Prediction}

\glsdisablehyper

\title{Lifelong Reinforcement Learning via Neuromodulation}
\author{
  Sebastian Lee$^{1}$, Samuel Liebana$^2$, Claudia Clopath$^1$, Will Dabney$^3$ \\\vspace{0.5em}
  $^1$Imperial College London \\ 
  $^2$University of Oxford \\ 
  $^3$Google DeepMind
}

\begin{document}

\begin{titlepage}
\thispagestyle{empty}
\maketitle

\begin{abstract}
Navigating multiple tasks---for instance in succession as in continual or lifelong learning, or in distributions as in meta or multi-task learning---requires some notion of adaptation. Evolution over timescales of millennia has imbued humans and other animals with highly effective adaptive learning and decision-making strategies. Central to these functions are so-called neuromodulatory systems. In this work we introduce an abstract framework for integrating theories and evidence from neuroscience and the cognitive sciences into the design of adaptive artificial reinforcement learning algorithms. We give a concrete instance of this framework built on literature surrounding the neuromodulators Acetylcholine (ACh) and Noradrenaline (NA), and empirically validate the effectiveness of the resulting adaptive algorithm in a non-stationary multi-armed bandit problem. We conclude with a theory-based experiment proposal providing an avenue to link our framework back to efforts in experimental neuroscience.
\end{abstract}

\end{titlepage}

\newgeometry{top=20mm,bottom=25mm,right=30mm,left=30mm}

\section{Introduction}

Initially catalysed by seminal work on the Atari-57 benchmark~\citep{mnih2015human}, recent research progress in \gls{rl} and deep RL has led to a string of impressive achievements in various domains including strategy games~\citep{silver2017mastering}, multi-agent games~\citep{berner2019dota}, as well more applied settings~\citep{bellemare2020autonomous, mirhoseini2020chip}. However, many of these methods rely on a combination of hyper-parameter tuning and various heuristics to achieve this success. For example, the $\epsilon$-greedy exploration strategy with a linear decay schedule for $\epsilon$ remains nearly ubiquitous in value-based \gls{rl}~\citep{hessel2018rainbow}. Other hyper-parameters such as the discount factor and learning rate are also typically chosen heuristically or optimised anew in each setting with expensive grid searches. As the focus of \gls{rl} research shifts from episodic single-task domains toward lifelong, multi-task, and continual learning problems, the reliance on per-domain tuning will be especially restrictive. Although recent work in meta-learning and meta-gradient~\gls{rl} have---in part motivated by this need---made significant progress to this end~\citep{finn2017model, xu2018meta, gupta2018meta}, these methods often come with their own additional hyper-parameters, rely on strong assumptions about the training regime, and/or involve high-variance gradient based updates~\citep{nagabandi2018learning, parker2022automated}.

In this paper we take a different approach, namely, we draw inspiration from an evolving understanding of the processes involved in animal learning to inform a novel framework for developing adaptive~\gls{rl} algorithms. We consider in particular the roles of various neuromodulators---chemicals that modulate neuronal responses in the brain---in allowing animals to explore, learn and interact effectively with changing environments. Our primary contributions in this regard can be summarised as follows: (1) a proposed framework for developing adaptive~\gls{rl} algorithms deeply rooted in fundamental neuroscience research, (2) an example instantiation of the framework to commonly used \gls{rl} hyper-parameters, (3) an empirical evaluation of the framework instance on a non-stationary multi-armed bandit task. 

In the latter part of the paper, we discuss the possibility of extending our framework for neuroscience experiment proposals, making use of the common ground it can establish between ideas in cognitive sciences and machine learning. To this end, we devise a more general framework proposing a closed-loop protocol to interconnect the two fields. Thus, the intended audience of this work includes both machine learning researchers interested in drawing from ideas in neuroscience, and neuroscientists seeking to use tools and formalism from machine learning. 

\section{Background}\label{sec: background}

\textbf{Reinforcement Learning.} The \gls{rl} problem is often formalised in terms of the interaction of an agent with a~\gls{mdp}. A~\gls{mdp} is defined by the tuple $\langle\mathcal{S},\mathcal{A}, P, R, \gamma, p_0\rangle$ with state space $\mathcal{S}$; action space $\mathcal{A}$; state transition probabilities $P: \mathcal{S}\times\mathcal{A} \to \Delta(\mathcal{S})$, which maps a state-action pair to a distribution over next-states;
reward function $R:\mathcal{S}\times\mathcal{A}\times\mathcal{S}\to\mathbb{R}$, which defines scalar reward given to agent from the environment at each step; discount factor $\gamma\in[0,1)$, which determines the trade-off between short and long-term rewards; and the initial state distribution $p_0 \in \Delta(\mathcal{S})$ (where $\Delta(X)$ denotes the space of probability distributions for set $X$). For some policy $\pi: \mathcal{S}\to \Delta(\mathcal{A})$, the expected sum of discounted rewards, known as the return, from a given state $s \in \mathcal{S}$, is given by the value function, $V^\pi(s) = \mathbb{E}[ \sum_{t=0}^\infty \gamma^t r_t | s_0 = s]$. The~\gls{rl} objective is to find a policy that maximises $V^\pi(s)$ in expectation over $p_0$. In~\glspl{mdp}~there always exists at least one such deterministic policy, known as the optimal policy and denoted $\pi^*$.

\textbf{Non-Stationarity.} Non-stationary processes are governed by probability distributions that change over time. We are implicitly concerned with the more general~\gls{pomdp}, where agent observations are conditioned on hidden states of an underlying~\gls{mdp}. Here the above tuple is augmented with $\Omega$---the set of observations, and $\mathcal{O}$---a conditional probability operator for each observation given the underlying state (and in some cases, action). In our work, the hidden state corresponds to the environment `context'; detection of---and subsequent adaptation to---new contexts is a crucial component of lifelong learning and one we argue is facilitated by neuromodulation in animals. Further effective non-stationarity may arise in~\gls{rl}, e.g. via shifting state and reward distributions encountered with non-stationary behavioural policies.

\textbf{Neuromodulation.} The foundational principles of~\gls{rl} have been heavily influenced by neuroscience and psychology. Richard Sutton's seminal doctoral work on temporal-difference learning \citep{sutton1978unified, sutton1988learning}, was explicitly inspired by Pavlovian conditioning. Since then ideas such as interleaved replay~\citep{lin1992self, mnih2013playing}, and intrinsic motivation~\citep{gopnik1999scientist, barto2013intrinsic} have also made their way into algorithmic innovations in~\gls{rl}. In this work we focus on a subset of the ties connecting these areas of decision-making, namely those associated with neuromodulation. As a functionally diverse set of systems, neuromodulation in the brain exerts global control of neural circuits, while simultaneously altering intrinsic excitability and synaptic strength at the level of individual neurons~\citep{katz1999beyond, marder2002cellular, froemke2015plasticity}. Neuromodulation has long been understood as an influential player in flexible cognition and learning and is thus a natural candidate to inspire research in lifelong~\gls{rl}.

\section{Neuromodulation for Adaptive RL}\label{sec: framework}

\begin{figure*}[t]
    \centering
    \includegraphics[width=0.95\linewidth]{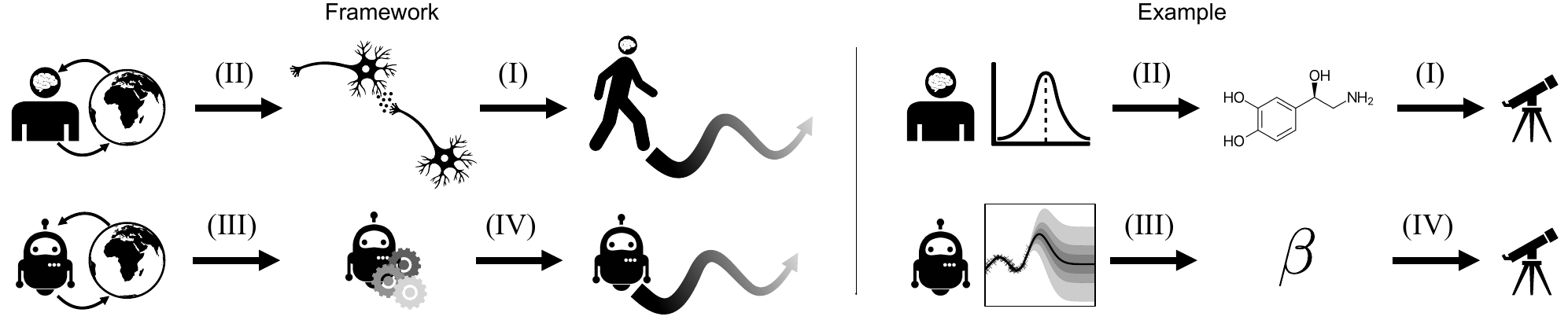}
    \caption{\textbf{Framework for using understanding of neuromodulatory systems to design adaptive RL algorithms.} \emph{left} The framework consists of four components: (I) a hypothesised connection between a neuromodulatory system and an RL algorithm hyper-parameter, i.e. what does the neuromodulator do?; (II) an understanding of what quantity the neuromodulator is believed to signal or measure; (III) an analogous quantity measurable as part of the artificial agent-environment interaction; (IV) a functional form mapping this estimate to a hyper-parameter value. \emph{right} An example, with details given in the main text, of the framework applied to (I) a hypothesised connection between Noradrenaline (NA) and exploratory behaviour; (II) where NA is believed to signal unexpected uncertainty; (III) which can be understood as epistemic uncertainty and estimated with ensemble methods; (IV) providing a functional mapping to the inverse temperature hyper-parameter to modulate the exploration-exploitation trade-off.} \label{fig:neuromodulat_rl_framework}
\end{figure*}

Here, we attempt to devise a suitable framework for integrating our current understanding of neuromodulators within mammalian brain systems into artificial~\gls{rl} algorithms in order to improve their adaptation and exploration capacities (see~\autoref{fig:neuromodulat_rl_framework} for framework outline). 

\subsection{What do Neuromodulatory Systems Do? (component I)}

The objective for the first component is to develop an informed hypothesis about a connection between the way a particular neuromodulatory system affects learning and behaviour in animals, and the way some hyper-parameter or other quantity affects learning and learned policies in an RL algorithm. Numerous mappings have been made between neuromodulatory systems and reinforcement learning algorithms~\citep{niv2009reinforcement, huertas2016role}. One of the earliest---and arguably most influential---of these mappings is due to~\citet{doya2002metalearning}, who proposed a very general framework for relating four key neuromodulators (dopamine, serotonin, acetylcholine, and noradrenaline) to quantities in~\gls{rl}. While providing a potentially overly simplistic model, it has shaped the way we think about~\gls{rl} and neuroscience, and lays the foundation for much of this work---especially our application of component (I). We briefly recapitulate his framework here along with other relevant evidence from the neuroscience literature. It should be emphasised that this mapping is~\emph{one}~choice for this component, and that our framework principally permits others; a broader discussion of this flexibility can be found in~\autoref{app: speculative}.

\textbf{Dopamine (DA).} The most well established connection between a neuromodulator and \gls{rl}, which follows decades of experimental evidence~\citep{schultz1993responses, schultz1997neural} and continues to inspire contemporary neuroscience research~\citep{dabney2020distributional, gardner2018rethinking, lerner2021dopamine, maes2020causal}, is the correspondence between the~\gls{rpe}~in~\gls{rl} and dopamine signals in various neurons of the mid-brain. In the~\gls{rpe} theory, dopamine activity is higher if reward is higher than predicted, remains at some baseline when reward does not deviate from prediction, and is suppressed when less reward is received than predicted. Note that when rewards are sufficiently spread out (or the discount factor is very low), as is often the case by construction in neuroscience, the~\gls{rpe} and TD-error are equivalent.

\textbf{Noradrenaline (NA).} NA has been implicated in arousal, relaxation, urgency and attention~\citep{avery2017neuromodulatory}. However the primary theories for noradranergic function fall into two categories: under the `adaptive gain theory' numerous studies have shown a connection between NA and sharpened neuronal responses via an increase in input-output gain~\citep{aston2005integrative, gilzenrat2002simplified, servan1990network, usher2002neuromodulation}; on the other hand the `network reset' theory suggests the noradranergic system is responsible for functional re-organisation of cortical activity under shifts in environmental contingencies, thus allowing for behavioural adaptation~\citep{bouret2005network}. In general, NA is highly likely to be involved in mediating between exploration and exploitation modes whereby high phasic and low tonic modes are associated with animal engagement (exploitation), and high tonic activity leads to distractable states (exploration)~\citep{aston1994locus}. Consequently,~\citet{doya2002metalearning}~proposes that NA may be responsible for tackling the exploration-exploitation trade-off via the inverse temperature of a softmax policy.

\textbf{Acetylcholine (ACh).} ACh has been shown to control the synaptic plasticity (propensity for weight change) in various parts of the brain~\citep{partridge2002nicotinic, rasmusson2000role, doya2008modulators}; recent findings of anti-correlation between ACh and DA release in the striatum indicates this could be achieved by gating the effects of dopamine~\cite{krok2023intrinsic, chantranupong2023dopamine}. ACh is thought to help control the balance between and rates of memory update and memory storage~\citep{hasselmo1993acetylcholine, hasselmo2004high, doya2008modulators}, for instance by enhancing sensory input over recurrent feedback~\citep{disney2007gain, goard2009basal, chen2015acetylcholine}. As a result of these studies,~\citet{doya2002metalearning} proposes an analogy from ACh to the learning rate. 

\textbf{Serotonin (5-HT).} Trading off immediate and long-term rewards is a crucial component to animal learning and the~\gls{rl} problem. The function of the serotonergic system is often studied alongside the dopaminergic system with oftentimes contradictory studies on whether they work together, independently or in opposition~\citep{daw2002opponent, graeff1997dual, fletcher1995effects, parsons1998serotonin1b, durstewitz2002computational, williams2002physiological, sershen2000serotonin}. These complex interactions, along with evidence linking serotonin with changes in delay discounting \citep{schweighofer2008low, tanaka2007serotonin, miyazaki2011activation}, lead \citet{doya2002metalearning} to propose serotonin as an analogue for the discount factor, $\gamma$.

In summary, \citet{doya2002metalearning} hypothesises mappings from dopamine, noradrenaline, acetylcholine, and serotonin to TD-error, an exploration parameter $\beta$ (inverse temperature), learning rate $\alpha$, and discount factor $\gamma$ respectively. This is shown for illustration purposes in the annotations of the Q-learning~\citep{watkin1993} update and softmax policy shown in~\autoref{fig: doya_labelled}.
\begin{figure}[h!]
\vspace{1cm}
\begin{equation}
    \label{eq:annotated_doya}
    Q(s,a)\leftarrow Q(s,a) + \tikzmarknode{lr}{\highlight{blue}{$\alpha$}}\underbrace{\Big[r \;+ \tikzmarknode{gamma}{\highlight{red}{$\gamma$}}\max_{a'} Q(s',a') - Q(s,a)\Big]} \quad \mathbb{P}(a_i|s) = \frac{e^{\tikzmarknode{beta1}{\highlight{black!30!green}{$\beta$}}Q(s, a_i)}}{\sum_{j=1}^{|\mathcal{A}|} e^{\tikzmarknode{beta2}{\highlight{black!30!green}{$\beta$}} Q(s, a_j)}}
    \vspace{0.8cm}
\end{equation}
\begin{tikzpicture}[overlay,remember picture,>=stealth,nodes={align=left,inner ysep=1pt},<-]
    \path (lr.north) ++ (0,1em) node[anchor=south east,color=blue!60] (scalep){\textbf{Learning rate $\leftrightarrow$}\\ Acetylcholine (ACh)};
    \draw [color=blue!60](lr.north) |- ([xshift=-0.3ex,color=blue]scalep.south west);
    
    \path (lr.south) ++ (-2,-2em) node[anchor=north west,color=brown] (mean){\textbf{TD-error $\leftrightarrow$} Dopamine (DA)};
    \draw [color=brown!57]([xshift=+17.4ex, yshift=-2.5ex]lr.south) |- ([xshift=-0.3ex,color=blue]mean.south west);
    
    \path (gamma.north) ++ (6,2em) node[anchor=south east,color=red!50] (scalep){\textbf{Discount factor $\leftrightarrow$} Serotonin (5-HT)};
    \draw [color=red!50](gamma.north) |- ([xshift=-0.3ex,color=red!50]scalep.south east);
    
    \path (beta1.south) ++ (-3.7,-2.5em) node[anchor=north west,color=black!30!green!50] (mean){\textbf{Exploration $\leftrightarrow$}\\ Noradrenaline (NA)};
    \draw [color=black!30!green!50](beta1.south) |- ([xshift=-0.3ex,color=black!30!green!50]mean.south west);
    \draw [color=black!30!green!50](beta2.south) |- ([yshift=-8ex,color=black!30!green!50]beta1.south);
\end{tikzpicture}
\vspace{-0.1cm}
\caption{$Q$-learning update and softmax policy labelled by the mapping between neuromodulators and ~\gls{rl}~hyperparameters proposed by~\citet{doya2002metalearning}.}\label{fig: doya_labelled}
\end{figure}

\subsection{What Do Neuromodulatory Systems Signal? (component II)}

The second component of the framework outlined in~\autoref{fig:neuromodulat_rl_framework} is inclusion of theories and evidence from neuroscience about the cause of various neuromodulatory responses; what do they signal or measure? We provide short discussions of this question for each aforementioned neuromodulator; a thorough review on neuromodulatory systems can be found in~\citet{avery2017neuromodulatory}.

\textbf{DA \& RPEs.} The hypothesis that dopamine encodes a temporal-difference error, or \gls{rpe}, is one of the most widely studied ideas in neuroscience, and has seen attention from theoretical and experimental communities alike. Despite its early success~\citep{schultz1997neural}, the simple picture of scalar~\gls{rpe} has been challenged by findings of high dopamine neuron diversity and implication of dopamine with an array of other functions including movement and motivation~\citep{lerner2021dopamine}. More sophisticated theories of dopaminergic functions have emerged to account for this more complex picture including Bayesian interpretation~\citep{gershman2019believing}, return distributions~\citep{dabney2020distributional}, reward shaping~\citep{akiti2021striatal}, and motivation / opportunity costs~\cite{niv2007tonic}. Nevertheless it remains widely understood that dopamine plays a key role in reward learning.

\textbf{NA/ACh \& Uncertainties.} In a series of papers, Yu and Dayan propose that ACh and NA signal \emph{expected} and \emph{unexpected} uncertainty respectively~\citep{dayan2006phasic, dayan2002expected, angela2005uncertainty}. They argue that interaction with stochastic environments that have dynamic (non-stationary) contexts requires multiple uncertainty types---in particular expected and unexpected uncertainty---for optimal inference. At a high level, expected uncertainty is the variability in an outcome given a certain context, while unexpected uncertainty is a surplus in variability due to a change in context. 
For example, we may construct probabilistic models for temperature in London conditional on the current season. A change in context---e.g. climate change or a natural disaster---may result in temperatures inconsistent with that model, resulting in unexpected uncertainty. 
While the experimental picture is far from complete, this theoretical model has empirical support. \citet{lawson2021computational} find that blocking NA during a probabilistic association task hampered performance under unexpectedness. \citet{iglesias2013hierarchical} show that cholinergic activity in the basal forebrain correlates with subject uncertainty about probabilistic outcomes following predictive cues; likewise~\citet{hangya2015central} show similar scaling with unexpectedness of reinforcement signals. More recently, following evidence that blocking ACh induces greater reliance on prior knowledge over new sensory information, ~\citet{marshall2016pharmacological} proposed another compelling theory for ACh, namely that it balances attribution of uncertainty within and between environmental contexts.

\textbf{Serotonin.} Serotonin has been widely implicated in a range of social behaviours and settings including stress, anxiety and motivation~\citep{buhot1997serotonin, kiser2012reciprocal}. Notably, various classes of drugs used to treat depression have the property of increasing synaptic serotonin concentrations by blocking re-absorption in the presynaptic neuron~\citep{masand1999selective}. Despite these findings and the attention afforded to them by neuroscience, psychology, and pharmacology communities, a normative or even computational theory
for the control of serotonin levels in response to environmental stimuli largely eludes us. Therefore, although it is a key neuromodulator and central to e.g.~\citet{doya2002metalearning}, there is insufficient convergent evidence on serotonin for a single suitable interpretation under our~\gls{rl} framework.

\subsection{Measuring Analogous Signals in Agent-Environment Interaction (component III)}
The third component of our proposed framework consists of identifying and measuring quantities in the~\gls{rl} agent-environment interaction that are analogous to the signals that various neuromodulators are hypothesised to signal in component (II). For the case of dopamine signalling a~\gls{rpe}, this consists simply of observing the reward and value function prediction. Noradrenaline and Acetylcholine~e.g.~under Yu and Dayan's theory, present a more intricate example where---under our framework---an~\gls{rl} agent would aim to measure notions of expected and unexpected uncertainties. These two quantities can be better understood as corresponding to \textit{aleatoric} and \textit{epistemic} uncertainties (see~\autoref{app: uncertainties}). The sub-problem of estimating these uncertainties has been studied extensively in \gls{rl} and machine learning more broadly \citep{kendall2017uncertainties,bellemare2017distributional,clements2019estimating, jain2021deup, hullermeier2021aleatoric}. For instance, one definition from~\citet{clements2019estimating} involves appropriate statistics of the posterior distribution over the parameters of the model and the quantiles of the return distribution:
\begin{align}\label{eq:clem_main}
    \mathcal{E}(s, a) &= \mathbb{E}_{i\sim\text{Unif}(1, N)}[\mathbb{V}_{\mathbf{\theta}\sim P(\mathbf{\theta}|D)}(y_i(\mathbf{\theta}; s, a))],\\
    \mathcal{A}(s, a) &= \mathbb{V}_{i\sim\text{Unif}(1, N)}[\mathbb{E}_{\mathbf{\theta}\sim P(\mathbf{\theta}|D)}(y_i(\mathbf{\theta}; s, a))];
\end{align}
where $\mathbb{E}$ and $\mathbb{V}$ are expectation and variance operators, $\text{Unif}$ is the uniform distribution, $\theta$ are the model parameters, $D$ is the data, $y_i$ is the $i^{\text{th}}$ estimated quantile, and $s, a$ are the state and action. Depending on modelling choices and the setting more broadly, these may be difficult to estimate; many methods resort to variants of ensemble methods~\cite{osband2016deep, jiang2022uncertainty}.

\subsection{Finding Functional Forms Back to Hyper-parameters (component IV)}
Once the appropriate measurements have been made, the final component of our framework entails constructing a function to translate these raw measurements into mathematically valid and behaviourally relevant values for the hyper-parameter (e.g. respecting bounds on the quantity), thereby bringing the mapping back full circle. As a first-order example, consider the following forms for learning rate and inverse temperature based upon $\mathcal{E}$ and $\mathcal{A}$:
\begin{equation}
    \alpha(s, a) = \frac{\mathcal{E}(s, a)}{\mathcal{E}(s, a) + \mathcal{A}(s, a)} \quad \beta(s) = \frac{1}{\langle\mathcal{E}(s, \hat{a})\rangle_{\hat{a}}},
    \label{eq:uncertainties}
\end{equation}
where $\langle\rangle_{\hat{a}}$ indicates an average over actions.
These are rooted in~\citet{marshall2016pharmacological} \&~\citet{doya2002metalearning} in relating the learning rate to uncertainty balance via ACh\footnote{Interestingly this functional form for the learning rate is reminiscent of the gain terms in the Kalman filtering literature~\citep{geist2010kalman}.}, and on~\citet{dayan2002expected}~\&~\citet{doya2002metalearning} in relating the inverse temperature to the unexpected uncertainty via NA. Crucially, this mapping is made with the~\emph{tonic} levels of NA~\citep{berridge2003locus, tervo2014behavioral}, rather than the high~\emph{phasic} modes in response to external stimuli on which Doya bases his arguments~\citep{aston1994locus}; hence we arrive at the inverse relationship.

\section{Doya-DaYu Agent}\label{sec: agent}

Having outlined the framework for using neuromodulation for~\gls{rl}~algorithm design above, we now introduce an~\emph{example} instance of the framework, which we call the~\emph{Doya-DaYu} agent. The first two components of our framework as employed in the agent are the following mappings:
\begin{itemize}[itemsep=1mm, parsep=0pt]
    \item Learning Rate:\\
    $\alpha$ $\leftrightarrow$ ACh $\leftrightarrow$ Uncertainty Balance\\ \citet{doya2002metalearning}~\&~\citet{marshall2016pharmacological}
    \item Inverse Temperature:\\
    $\beta^{-1}$ $\leftrightarrow$ NA $\leftrightarrow$ Unexpected Uncertainty\\ \citet{doya2002metalearning}~\&~\citet{dayan2002expected}
\end{itemize}
The dopamine mappings discussed above are also implicit in the temporal difference methods we use. Components (III) and (IV) of the Doya-DaYu agent differ depending on the experimental setting.
In the tabular case, which we evaluate in a bandit task (see~\autoref{fig: bandits}), we train an ensemble of agents. For the expected (aleatoric) uncertainty, we directly estimate the variance of the return distribution $\text{Var}(G(a))$ via~\citep{white1988mean}:
\begin{equation}
    \text{Var}(G(a)) \leftarrow \text{Var}(G(a)) + \alpha_{G}[\delta^2 - \text{Var}(G(a))],
\end{equation}
which corresponds to the reliability index~\citep{sakaguchi2004reliability} under stationary environments. We take the variance over the ensemble of the mean values to be the unexpected (epistemic) uncertainty. Further options for estimating these quantities are discussed in~\autoref{app: tabular_uncertainty_methods}. For component (IV), we use the functional forms given in~\autoref{eq:uncertainties}.

\subsection{Multi-Armed Bandit Experiment}\label{sec: experiments}

\begin{figure}
    \centering
    \frame{\includegraphics[width=\textwidth]{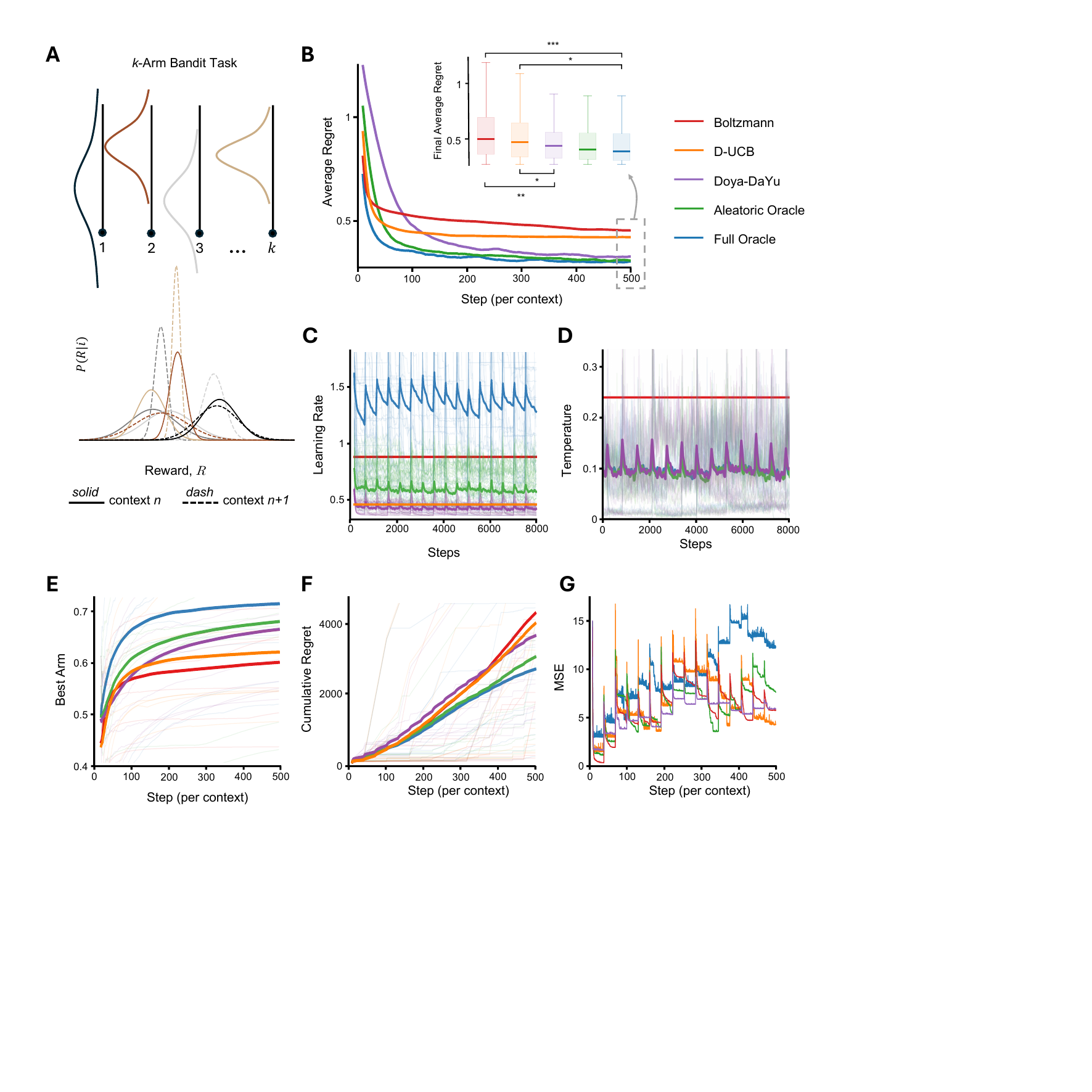}}
    \caption{\textbf{Multi-Armed Bandit Evaluation}. 210 seeds; $k=5$, $p=0.4$, $M=500$, $\mu_{\min}=-5$, $\mu_{\max}$=5, $\sigma_{\min}=0.001$, $\sigma_{\max}=2$. Dark lines show the mean; background traces show a random subset of individual seeds. \textbf{A} (top) schematic of the $k$-armed bandit: the agent must choose from one of $k$ arms with normally distributed payouts. (bottom) Example payout distributions in our multi-armed bandit task. Each colour represents a different arm. Solid/dashed lines show distributions before/after a context switch. Agents must model these context switches and explore to find the new high payout arm, while also navigating changing payout variances. \textbf{B} Average regret per context; while regret is initially higher in new contexts for Doya-DaYu, its final performance is superior. \emph{(inset)} final average regret box plot; braces marked *, **, *** indicate $p<0.1$, $p<0.01$, and $p<0.001$ estimated via an initial one-way ANOVA followed by a Tukey HSD test. \textbf{C}, \textbf{D} Learning rates and temperatures adapt to spike at context switches and then decay. \textbf{E} Fraction of arm pulls that are optimal averaged over all context switches. On average, the full oracle Doya-DaYu agent selects the best arm significantly more than baseline models by the end of contexts.
    \textbf{F} Cumulative regret is lower for Doya-DaYu agents than Discounted-UCB and Boltzmann policies, especially as the number of context switches increases, and the space of payoff distributions encountered widens.~\textbf{G} Mean squared errors for value estimates over the course of learning. Different levels of exploration lead to different distributions of errors over the arms; having low estimation error on sub-optimal arms is not directly beneficial. It does however suggest that baseline algorithms are over-exploring in this task, while the full oracle has high average MSE over all arms. Further results, details of implementations including procedures for hyper-parameter optimisation in the baselines can be found in~\autoref{app: exp_details}}
    \label{fig: bandits}
\end{figure}

We construct a non-stationary multi-armed bandit task in which to evaluate the tabular Doya-DaYu agent. Each bandit has $k$ arms, and learning occurs over $N$ contexts. Each context (or `block') lasts $M$ steps, after which a switch occurs with probability $p$ (context block lengths vary). Contexts are determined by the $k$ payout distributions, all of which are Gaussians with means sampled uniformly between $\mu_{\min}$ and $\mu_{\max}$, and standard deviations sampled uniformly between $\sigma_{\min}$ and $\sigma_{\max}$. We compare Doya-DaYu against Discounted-UCB~\citep{garivier2008upper} and Boltzmann policies in~\autoref{fig: bandits}. Discounted-UCB aims to solve the exploration-exploitation trade-off by adding a term that scales with the estimator variance to the value estimates, thereby encouraging selection of arms with higher uncertainty. Although algorithms like Discounted-UCB are known to have logarithmic regret in many settings~\citep{auer2002finite}, this test-bed is significantly more involved (as indicated by the high variance results) with more environment contingencies and fewer restrictions e.g. on payoff range or number of context switches. As such there is added benefit from the adaptive capabilities of the Doya-DaYu agent.
We also include two oracle versions of the Doya-DaYu agent: the `aleatoric oracle', where the true aleatoric uncertainty $\mathcal{A}^*$ (arm variance) is given to the agent; and a `full oracle', where both $\mathcal{A}^*$ and the epistemic uncertainty $\mathcal{E}^*$ are provided. Here we follow definitions given by~\citet{lahlou2021deup} for the oracle epistemic uncertainty, where
\begin{equation}
    \mathcal{E}^*(s, a) = \mathcal{U}(s, a) - \mathcal{A}^*(s, a),
\end{equation}
where $\mathcal{U}$ is the total uncertainty. As such an `oracle' epistemic uncertainty can be computed from a large enough sample of the error and the oracle aleatoric uncertainty. In~\autoref{fig: bandits}, agents with more oracle information perform better---further giving credence to the underlying framework. With more sophisticated uncertainty estimation methods, these gaps should reduce.

\section{Feedback to Neuroscience Experiments}\label{sec: proposal}

In sections 2-4, we established a protocol and demonstrated the viability of using ideas from neuroscience and psychology to inspire agent design in~\gls{rl}. Armed with this framework,
we now ask how it in turn could be used to motivate experimental design and analysis in neuroscience. In doing so we follow recent calls for more theory-driven approaches to experimental neuroscience~\citep{biswas2022geometric, saxe2021if}. Successful examples of such endeavours outside~\gls{rl} include predictive coding in electric fish~\citep{muller2019continual, muller2022biophysical}, fly head direction cells~\citep{kim2017ring}, continuous attractor networks for grid cell activity on toroidal manifolds~\citep{gardner2022toroidal}, as well as other theories of structured learning in hippocampus and enthorinal cortex~\citep{whittington2020tolman}.

\begin{figure}[h]
    \centering
	\includegraphics[width=0.75\textwidth]{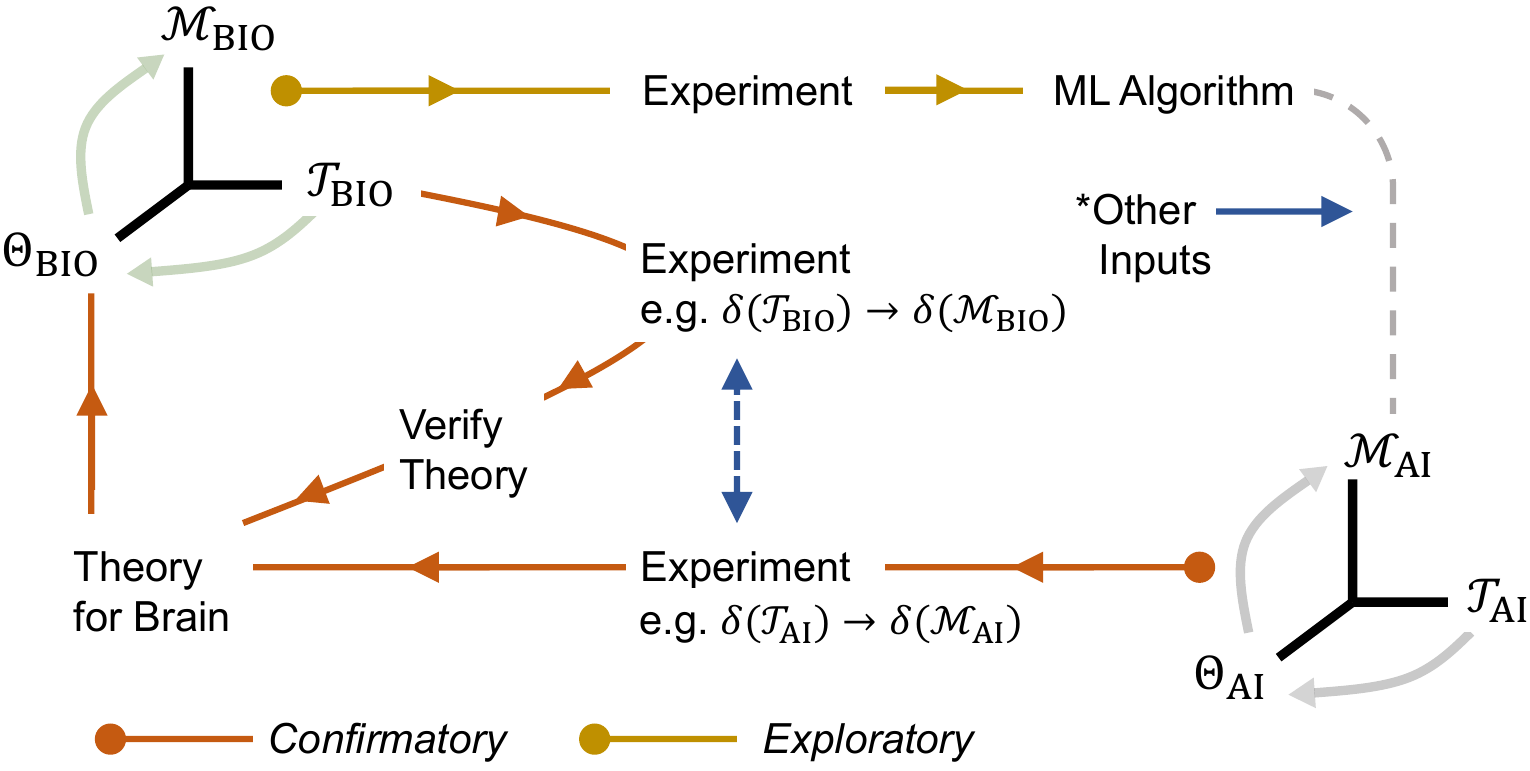}
    \caption{\textbf{Neuromodulation Framework as an  Experimental Design Paradigm.} The~\emph{exploratory} branch (starting in the top-left) where machine learning algorithms are proposed on the basis of experimental observations along ($\mathcal{T}_{\text{BIO}}$, $\Theta_{\text{BIO}}$, $\mathcal{M}_{\text{BIO}}$) axes. The~\emph{confirmatory} branch (starting in the bottom-right) is a theory-based experimental pipeline in which a model of animal learning, $\Theta_{\text{AI}}$, gives rise to a theory of the brain (see~\autoref{app: other_inputs} for discussion of `other inputs'). Theory can be verified by mapping experiments from $\text{AI}$ to $\text{BIO}$ axes via our neuromodulatory framework.}
    \label{fig: exp_proposal_abstract}
\end{figure}

Our central proposal, illustrated in~\autoref{fig: exp_proposal_abstract}, contains two branches. The first, which we call the~\emph{exploratory} branch of experiments (yellow), consists of proposals aimed at uncovering brain functions that could aid development of machine learning algorithms. Such efforts would be closely aligned with the philosophy of the algorithmic framework we presented earlier. The second branch, which we call ~\emph{confirmatory} (red), makes use of artificial models of learning to make concrete predictions about animal behaviour.

\usetikzlibrary{decorations.markings}
\tikzset{->-/.style={decoration={
  markings,
  mark=at position #1 with {\arrow {latex}}},postaction={decorate}}}

\definecolor{explore}{HTML}{C1666B}
\definecolor{confirm}{HTML}{D4B483}
\definecolor{autre}{HTML}{4281A4}

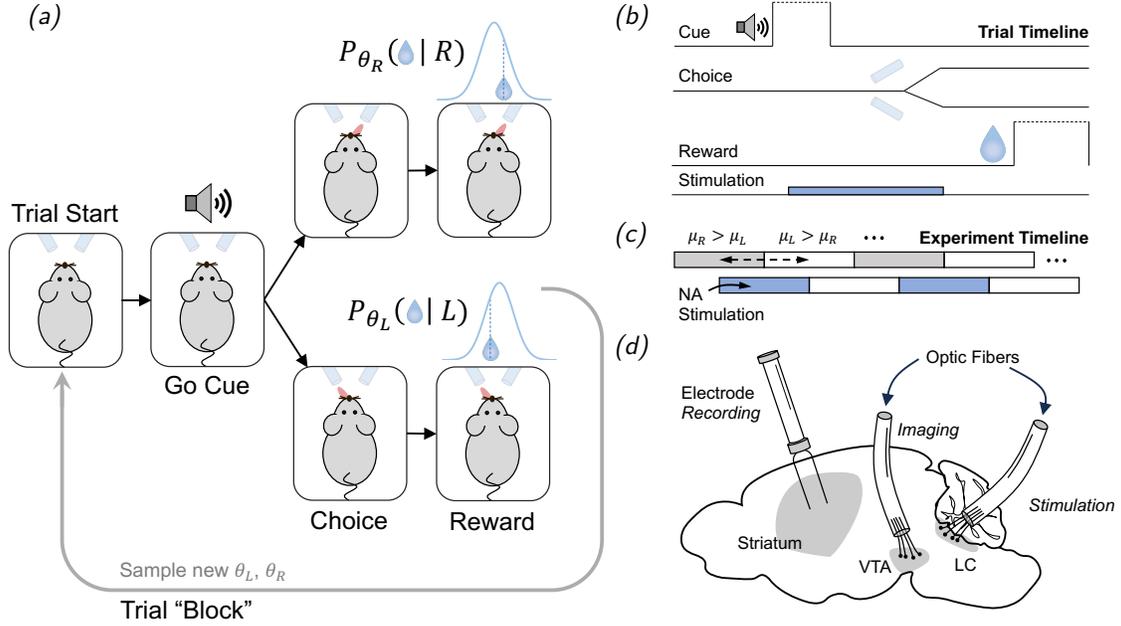
\begin{figure*}[t]
    \centering
    \begin{tikzpicture}
    \node [opacity=1, anchor=south west] at (0, 0) {\includegraphics[width=0.95\textwidth]{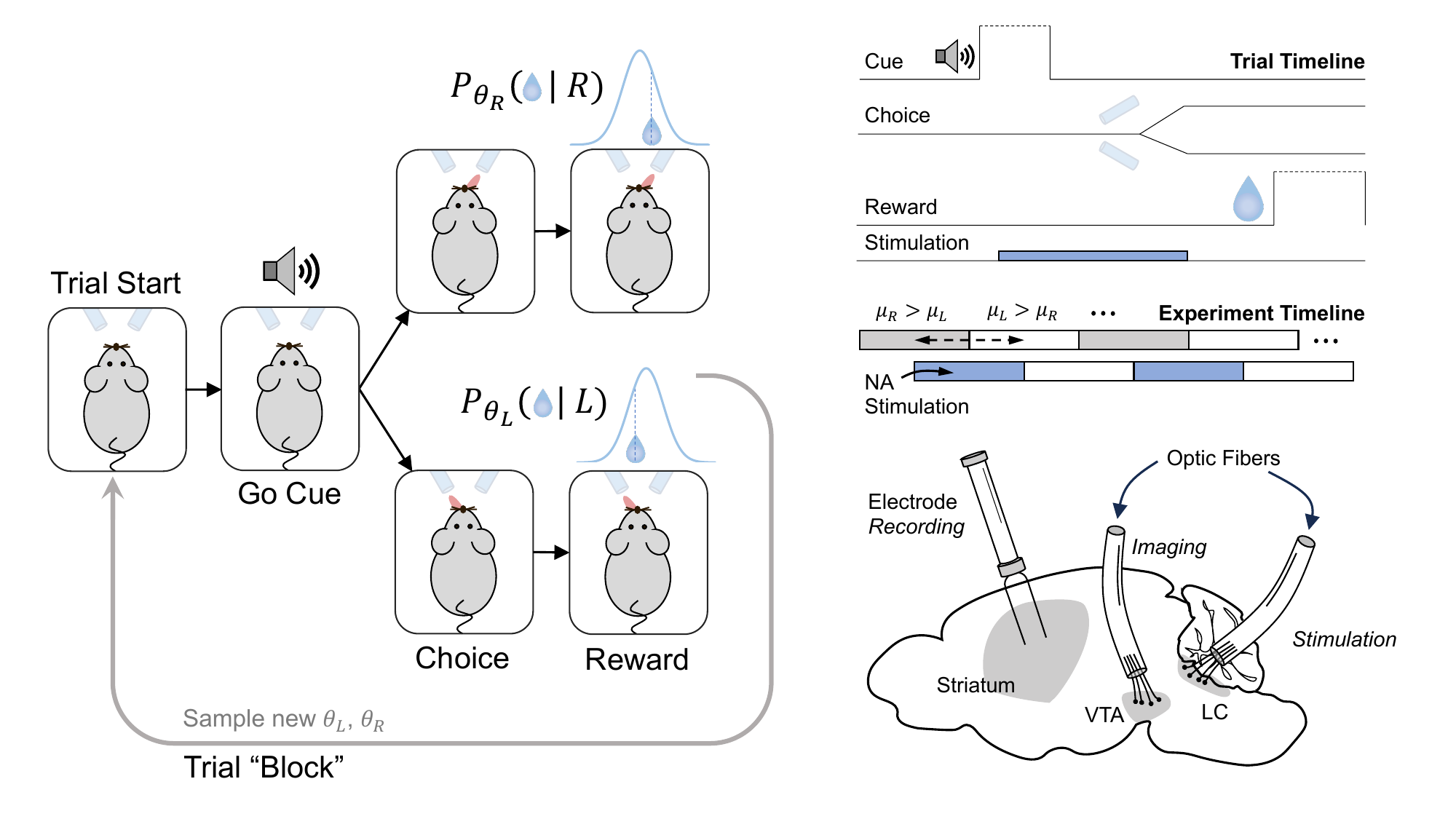}};
    \node [anchor=west] at (0.3, 8.2) {\emph{(a)}};
    \node [anchor=west] at (8.1, 8.2) {\emph{(b)}};
    \node [anchor=west] at (8.1, 5.3) {\emph{(c)}};
    \node [anchor=west] at (8.1, 3.8) {\emph{(d)}};
    \begin{subfigure}[b]{\textwidth}
    \phantomsubcaption
	\label{fig:4a}
	\end{subfigure}
    \begin{subfigure}[b]{0pt}
    \phantomsubcaption
	\label{fig:4b}
	\end{subfigure}
    \begin{subfigure}[b]{0pt}
    \phantomsubcaption
	\label{fig:4c}
	\end{subfigure}
    \begin{subfigure}[b]{0pt}
    \phantomsubcaption
	\label{fig:4d}
	\end{subfigure}
    \end{tikzpicture}
    \vspace{-0.5em}
    \caption{\textbf{Concrete Experimental Proposal} \emph{(a)} Proposed task to mirror our multi-armed bandit task in~\autoref{sec: experiments}; on each trial---following a `go' cue---mice lick from one of two ports, after which they are given water reward sampled from some underlying distribution for the port selected. Upon reaching proficiency in selecting the more rewarded port, a new trial `block' with new distributions begins. This paradigm is similar to those used e.g. by~\citet{duan2021cortico} and~\citet{sabatini_two_armed}. \emph{(b)} \textit{Experiment timeline}: the experiment is divided into blocks of trials where one of the two arms has a higher expected reward $\mu$. The stimulation protocol is designed such that stimulation epochs occur through a reversal period, and are counter-balanced across side reversals. \emph{(c)} \textit{Trial timeline}: Phasic stimulation is applied from `go' cue to after port choice to probe the influence of manipulating NA levels on behaviour. \emph{(d)} \textit{Neural stimulation/recording schematic}: An optic fiber over Locus Coeruleus (LC) enables optical inhibition/excitation of NA neurons; the fiber over Ventral Tegmental Area (VTA) allows for confirming changes in NA concentration upon laser stimulation; and the electrode in striatum, believed to encode action values for decision-making, enables recording of striatal activity throughout the task.}
    \label{fig: exp_proposal}
\end{figure*}

\textbf{Experiment Axes.} $\mathcal{T}$, $\Theta$ and $\mathcal{M}$ can vary in a given experiment. $\mathcal{T}$ represents experimental conditions or task, for instance likelihood of context change in the bandit; $\Theta$ represents the agent (artificial or biological), for instance levels of a given neuromodulator or model parameter values; and $\mathcal{M}$ represents experimental measurements, including recording or imaging data, as well as performance or behaviour metrics. Neuroscience experiments generally consist of changing some condition for the animal; this can be done either directly by perturbing along the $\Theta$ axis, or by modifying the environmental conditions on $\mathcal{T}$. By observing neural and behavioural changes, it is possible to form theories and make conclusions about the brain. The axes describing biological experiments can be related to the equivalent axes describing experiments using models of learning via the framework from~\autoref{sec: framework}. Thus we can hope to use the conclusions from neuroscience to inform machine learning algorithms. This is not a new insight as such, although we argue that there are few concrete frameworks for facilitating this transfer of knowledge---especially with focus specifically on neuromodulation. Arguably more illuminating is the opposite direction: from models of learning, to theories---and subsequently experiments---of the brain, which we encapsulate in the~\emph{confirmatory} branch. Here the assumption is that we can take an artificial model of learning and treat it as a model of some brain function. We can perform experiments with this model and in some cases tractably analyse the model (e.g. regret bounds in bandits~\citep{auer2002finite}, statistical and more recently deep learning theory~\citep{vapnik1999overview, saxe2013exact}). Probing the model can allow us to develop theories for the brain, which we can test by constructing experiments on the equivalent $(\mathcal{T}, \Theta, \mathcal{M})$ axis in the animal.

\textbf{Confirmatory Proposal.} In~\autoref{fig: exp_proposal} we present a concrete example of an experiment proposal along the~\emph{confirmatory} branch with the high-level objective of confirming the suitability of Doya-DaYu as a model of animal learning. In particular, we test whether the temperature $\beta^{-1}$ is implemented through noradrenaline (NA). To probe this, we designed a two-armed bandit task for mice (\autoref{fig:4a}) in which they decide between one of two lick ports after an auditory `go' cue (\autoref{fig:4b}). The lick ports deliver water reward stochastically following fixed reward magnitude (e.g., 3$\mu l$) Bernoulli distributions, with success probabilities summing to 1 across arms. The task is divided into blocks of trials (\autoref{fig:4c}), 
where one of the two ports has a higher average reward than the other ($\mu_L$ vs $\mu_R$). When the mouse stably exceeds a certain threshold of proficiency in one block (higher probability of choosing high reward port), the reward contingencies are reversed such that the low reward port becomes the high reward port and vice versa. The reversals ensure that the animals remain engaged in the task, avoid excessive habitual or `sticky' behaviour and allow repeated sampling of the stage where the animals learn the reward contingencies. To stop the mice from predicting reversal times, we sample a number of trials from an appropriate distribution to determine how long proficient behaviour is to be maintained before a reversal.

From the neural perspective, we investigate the causal effects of manipulating NA on choice behaviour. The mechanics of the Doya-DaYu agent put forward the hypothesis that lower (higher) levels of phasic NA lead to more (less) random behaviour by increasing (decreasing) the temperature of the softmax function. To test this hypothesis, we propose the phasic stimulation of NA neurons in the locus coeruleus (LC) using both excitatory and inhibitory opsins~\citep{carter2010opto} at the time of the `go' cue (\autoref{fig:4d}). We titrate the power and duration of laser stimulation to ensure physiological levels of NA manipulation by combining stimulation with fiber photometry recordings of NA release using GRAB sensors~\citep{feng2019grabNA} in known LC projection targets such as the ventral tegmental area (VTA)~\citep{schmidt2020}. Laser stimulation could also be paired with electrophysiological recordings, e.g., neuropixel~\citep{jun2017}, of key decision-making circuits such as striatum~\citep{doya2005,nomura2020optorecord}.

The stimulation protocol is depicted in~\autoref{fig:4c}. On a subset of blocks, the protocol involves stimulating before and after a reversal---more specifically, from just after proficiency pre-reversal, to just after proficiency post-reversal (with blocks counterbalanced across reversal types $\mu_R > \mu_L \leftrightarrow \mu_L > \mu_R$). This enables the collection of data from all stages of behaviour (learning, proficiency, reversal) with and without stimulation. This protocol also ensures that stimulation is not interfering with the typical function of learning and decision-making circuits by e.g. corrupting value information or impairing attention, as mice have to re-learn the contingencies before another reversal occurs. Our analysis would then focus on comparing the randomness (or entropy) of choice behaviour in epochs with and without stimulation. A sketch of our expected results and a complementary discussion can be found in~\autoref{app: exp_proposal_results}. Further details on the experiment design can be found in~\autoref{app:exp_proposal_full}. 

\section{Discussion}
In this work we have introduced an abstract framework for embedding ideas (theoretical and/or empirical) from neuroscience and psychology, in particular neuromodulatory control, into agent design in artificial reinforcement learning. In addition we have discussed a reverse framework in which models of learning can be used to inspire experimental choices in neuroscience. On the~\gls{rl} side, our work fits broadly into the category of lifelong or continual reinforcement learning. As discussed by~\citet{khetarpal2020towards}, this is related or could even be argued to subsume other paradigms including meta-learning, domain adaptation and transfer learning. Under their taxonomy, our framework sits under so-called learning-to-learn, which includes sub-problems like context detection, learning to adapt and learning to explore. Unlike many previous approaches~\citep{schmidhuber1990making,pathak2017curiosity, zintgraf2019varibad, kamienny2020learning}, the adaptive algorithms from our framework do not require alternative training regimes or differentiation of the loss with respect to hyper-parameters; rather the inductive biases for adaptation are drawn from neuroscientific principles.

In presenting various aspects of our framework we have made frequent reference to ideas in theoretical neuroscience, most prominently the seminal work by~\citet{doya2002metalearning}. In the experimental literature, there are numerous examples of protocols designed to study questions around neuromodulation with similar flavour to our proposal in~\autoref{fig: exp_proposal}~\citep{grossman2022serotonin, luksys2009stress, cohen2015serotonergic}. However such approaches tend to follow the~\emph{exploratory} branch as we have defined it, while we hope continued discourse on correspondence between artificial models of learning and biological agents can foster complementary approaches on the~\emph{confirmatory} path. Further to the neuromodulation framework discussed in this paper, future work could focus on exploring and confirming hypotheses around the function of neural codes in learning and decision-making under an analogous framework. Past examples are the posited representation of stimulus subjective value in prefrontal cortex~\citep{lak2020dopaminergic}, action value encoding in striatum~\citep{doya2005}, and evidence integration processes in parietal cortical regions~\citep{shadlen2001neural}. Such results could be combined with confirmed neuromodulatory mechanisms to construct more elaborate and insightful models of biological behaviour.

Neuroscience has a long tradition of influencing progress in AI. This has been renewed in recent years as connections between neuroscience and deep learning in particular are being explored~\citep{hassabis2017neuroscience}. The~\emph{Doya-DaYu} agent can in principle also be extended beyond the tabular setting and used in combination with a deep~\gls{rl} agent. We propose an agent in this direction, along with a navigational environment in which it could be evaluated in~\autoref{app: dist_rl}. For continual learning specifically, numerous works have proposed algorithmic innovations modelled on specific neural mechanisms~\citep{kaplanis2018continual}. Many of the examples we have used to illustrate our framework, e.g. using the reliability index for exploration policies or combining ideas from~\citet{doya2002metalearning} and~\citet{angela2005uncertainty}, are similar in flavour to suggestions that have been made elsewhere; for instance by~\citet{tanaka2007serotonin} or~\citet{modirshanechi2021surprise}---who also provide a comprehensive overview of surprise measures in artificial and biological learning. However to the best of our knowledge our framework is the first to propose a high-level abstract framework for using understanding of neuromodulatory mechanisms to design~\gls{rl} agents.

In the coming years we anticipate the machine learning community to place a greater emphasis on flexible and adaptive reinforcement learning algorithms. In parallel, with widespread developments in recording techniques across species, the behavioural influences of neuromodulation continue to drive the expansion of existing theoretical models within the field of neuroscience. We hope that alongside these research trajectories, the communities of neuroscience and machine learning continue to exchange ideas~\cite{marblestone2016toward, richards2019deep, botvinick2020deep}, and that our frameworks can provide one basis for this kind of cross-pollination. Although the frameworks are self-contained, in future we plan to build on this work with further discussion, example implementations and empirical evaluations.

\bibliography{main}
\bibliographystyle{unsrtnat}

\appendix
\section{Proposals for Framework}\label{app: speculative}

In the main text we focused on the four primary mammalian neuromodulators, and their relation to arguably the most ubiquitous hyper-parameters in~\gls{rl}. However the frameworks that we present can be applied in many other contexts; below we speculate on some of these potential avenues:

\textbf{Regularisation.} Numerous methods for regularisation have been developed in machine learning. These include dropout and other noise injection techniques, weight decay, penalty terms e.g. L1 or L2, and even data-augmentation and early-stopping; each of which involves its own hyper-parameter(s). While many of these are used primarily with the aim of preventing over-fitting, regularisation terms also appear in other contexts; for instance consolidation terms in continual learning, or consistency regularisers in semi-supervised learning. Clearly animals face some of the above problems too (e.g. around generalisation, memory consolidation), as well as additional problems around metabolic constraints. It is plausible that solutions in the brain involve neuromodulatory functions. For instance, the serotonergic system is a good candidate as an implementer of regularisation, due to its heightened activity during learning and salient events and its proposed function as a promoter of the long-term depression of synapses~\citep{wert2022dopamine}. 

\textbf{Architecture \& Connectivity.} Different neural network architectures are used in various parts of machine learning from simple feed-forward networks, to convolutional networks (including specification of filters, padding, pooling etc.), and recurrent networks (including specification of gating functions, hidden state properties etc.); all with a zoo of activation functions. Beyond this, other `tricks' include skip-connections or residual layers, FiLM layers, attention mechanisms, pruning methods etc. Questions around connectivity and modularity are also central to systems neuroscience, and neuromodulation is likely to be involved---especially in the development of these structures~\citep{faust2021mechanisms, foehring1999neuromodulation}.

\textbf{Replay \& Planning.} Across neuroscience and machine learning, methods around replay, planning and other cognate ideas are very prevalent. In~\gls{rl} alone, we often must consider trade-offs between model-based v.s. model-free learning, offline v.s. online learning, and replay buffers and prioritisation/scheduling. From a neuroscience perspective, these dichotomies arise along the lines of short-term v.s. long term memories \& hippocampal v.s. cortical processing, and neuromodulation has once again been proposed to mediate these dynamics~\citep{atherton2015memory, carr2011hippocampal}.

\textbf{Plasticity.} Beyond the learning rate, additional parameters govern weight changes in different contexts across machine learning. More sophisticated optimisers such as ADAM will have momentum and scheduled annealing parameters, meta-learning algorithms frequently have learning rates at different scales (i.e. an inner and outer loop learning rate) etc. More broadly trust regions in policy optimisation or other restrictions on distribution shifts in other contexts can also be understood as controlling plasticity (these could also be discussed in the context of regularisation), as can target networks for learning stability and trade-offs between Monte-Carlo and temporal-difference learning e.g. via eligibility traces. We have discussed dopamine in some detail in the main text, focusing primarily on the simple RPE theory. Over the years, various works have revealed a greater diversity of dopamine responses and function than initially put forward by the RPE theory~\citep{starkweather2021dopamine, lee2022vector, jeong2022mesolimbic}. Some of these functions may be promising candidates for use under our framework in relation to more complex elements of plasticity control outlined here.
\section{Epistemic \& Aleatoric Uncertainty}\label{app: uncertainties}

Broadly, epistemic uncertainty is a reflection of lack of knowledge; it is considered~\emph{reducible} since it can be improved with more experience or data. On the other hand aleatoric uncertainty is the inherent uncertainty in the task or environment; it is~\emph{irreducible} since there is a fundamental lower bound dictated by the environment, regardless of the quality of the model. Although these are well established notions, they can be formulated mathematically in various---often context dependent---ways. We covered two such definitions above: for general classes of predictors in supervised learning,~\citet{jain2021deup} define aleatoric uncertainty as the total uncertainty of a Bayes optimal predictor:
\begin{equation}
    \mathcal{A}(x) = \mathcal{U}(f^*, x),\label{eq: ale}
\end{equation}
where $f^*$ is the Bayes optimal predictor, meaning the best predictor in the model class. The epistemic uncertainty of any predictor $\hat{f}$ is then defined as the total uncertainty minus the aleatoric uncertainty, i.e.
\begin{equation}
    \mathcal{E}(\hat{f}, x) = \mathcal{U}(\hat{f}, x) - \mathcal{A}(x).
\end{equation}

For interactive learning and~\gls{rl},~\citet{clements2019estimating} define epistemic and aleatoric uncertainty in terms of statistics of the posterior distribution over the parameters
of the model and the quantiles of the return distribution (\autoref{eq:clem_main}).
Here, we are particularly concerned with the correspondence between epistemic~\&~aleatoric uncertainty, which are common parlance in the statistics and machine learning communities, and expected~\&~unexpected uncertainty, which are frequently used in theoretical neuroscience. As has been mentioned in various works, perhaps most notably by~\citet{yu2011uncertainty}, there is a `rough analogy' between expected uncertainty and aleatory on the one hand, and between unexpected uncertainty and epistemic uncertainty on the other. There is further analogy to be made to external disposition and internal ignorance from the behavioural economics literature. While many of these concepts are very closely related, the lack of precise mathematical terminology---particularly outside the statistics community---can leave space for ambiguity. For instance aleatoric uncertainty (e.g. as defined in~\autoref{eq: ale}) is the uncertainty defined with respect to an optimal predictor; it is determined by the environment and the model class. However in the common usage of the term `expected uncertainty', there is still an implication that the current state of knowledge of the agent is relevant, i.e. that it has not necessarily `converged' on an estimate. Stated differently, when it comes to expected and unexpected uncertainties there is a notion that within a context our attribution of uncertainty can move from the unexpected to the expected (c.f.~\citep{marshall2016pharmacological}); whereas in the statistical sense aleatoric uncertainty is very precisely defined and everything else is epistemic uncertainty. Nonetheless, for simplicity we make a direct connection between these terms in our work. Further discussion on this topic can be found in~\citet{yu2011uncertainty}, while a more rigorous mathematical treatment classifying these various measures can be found in~\citet{modirshanechi2021surprise}.

\section{Uncertainty Estimation Methods}\label{app: tabular_uncertainty_methods}

In~\autoref{sec: agent} we presented an agent with adaptive learning rate and temperature built around the neuromodulators acetylcholine and noradrenaline, and their purported relations to expected and unexpected uncertainty. In doing so we give simple prescriptions for how these uncertainties could be measured for a bandit task using an ensemble. There are however myriad options for performing these estimates from decades of literature in statistics and computer science, particularly for online learning and bandits. For instance the notion of unexpected uncertainty is closely related to the idea of change point detection, which has been frequently used in non-stationary bandits~\citep{aminikhanghahi2017survey}.

\section{Experimental Details}\label{app: exp_details}

In addition to the softmax baseline that chooses actions according to the distribution given in~\autoref{fig: doya_labelled}, we also present Discounted-UCB in our bandit ablation. The vanilla Upper Confidence Bound (UCB) algorithm selects actions by choosing preferentially both on the basis of current value estimates, and the variances of those estimates. In general for algorithms in the UCB family, the selection operation is given by:
\begin{equation}
    A(t) = \arg\max_k{\left[\hat{Q}_k(t, \gamma) + c_k(t, \gamma)\right]},
\end{equation}
where $A(t)$ is the action selected at time $t$, $k$ is the arm index, $\gamma$ is a discount parameter (set to 1 for vanilla case), and $c_k$ represents the confidence bound term. For Discounted-UCB:
\begin{equation}
    c_k(t, \gamma) = \xi\sqrt{\frac{\log(n_t(\gamma))}{N_k(t, \gamma)}} \quad N_k(t, \gamma) = \sum_{s=1}^t\gamma^{t-s}\mathds{1}\{\pi_s=k\},
\end{equation}
where $n_t(\gamma)=\sum_{k=1}^KN_k(t,\gamma)$. The idea behind this discounting is that in non-stationary environments we want to exponentially down-weight counts of actions in the past. Further details on this algorithm, and it's variants can be found in~\citet{garivier2011upper}.

For both baselines, we perform a simple grid search to select the hyper-parameters used in the plot in~\autoref{fig: bandits}. For the Boltzmann baseline, the grid comprised 0.05 increments from 0 to 1 for the learning rate and inverse temperature. For Discounted-UCB we searched over $\gamma=\{0.9, 0.99, 0.999, 0.9999\}$, 0.1 increments from 0.5 to 1.5 for $\xi$, and 0.05 increments from 0.1 to 1 for the learning rate. The criterion was simply the final cumulative regret. For the Boltzmann policy, this gave 0.25 for both the learning rate and inverse temperature; for the Discounted-UCB it gave $\gamma=0.9999$, and 0.25 for the learning rate. 

We did not conduct any additional grid searches for the Doya-DaYu agents (for instance over the ensemble size or tweaking the functional forms by other constants), so one can expect extra performance could be gained.

\section{Models of the Brain}\label{app: other_inputs}

Models and theories of the brain such as the Doya-Dayu agent discussed in this paper are rarely the product of solely experimental observations from brain studies. AI agents typically have a wealth of features that derive from other fields such as mathematics, physics or computer science which allow the design of functional systems for diverse purposes. So while it is possible to take models of learning inspired by the brain from the exploratory branch and feed them back into the confirmatory branch, it will often be beneficial to incorporate these additional features. We denote these as `Other Inputs' in our two-branch neuro-AI framework shown in~\autoref{fig: exp_proposal_abstract}, and will unpack what we mean by this below.

Experimental studies in the brain sciences, although extremely insightful and inspirational for researchers in AI, are very limited in what they can observe and manipulate in the brain. This means that it is unrealistic to believe that, at least with current techniques, we are anywhere close to creating an AI system that perfectly reproduces the mechanistic principles on which the brain operates. The brain is extremely complex, and in addition to its independent workings, has a wealth of interactions with the rest of the body which are crucial and yet remain highly understudied. Consequently, we tend to resort to uncovering high-level principles that govern the working of parts of the brain and derive inspiration from them to build systems capable of emulating certain aspects of those regions. Classic examples are those of the hippocampus and working memory, dopamine and reward prediction error, cortex and connectionist models etc. These principles are typically embedded into theoretical frameworks that have been designed to operate on existing technology and implement mechanisms that can be useful to solve everyday problems. Importantly, these additions from independent lines of research prove crucial for the development and refinement of our theories of the brain, as they provide practical contexts and theoretical frameworks that give new perspectives on brain function, as well as fully characterised systems with emergent properties that can help us better understand similar ones present in the biological brain.

Another important source of other inputs are the different constraints under which the brain and artificial agents operate. Evolution has shaped and adapted the brain such that it performs optimally under the conditions imposed by its biological substrate. These are not necessarily the same constraints as those under which artificial systems operate. However, these differences encourage discussions about optimality which prove very useful to further research in both domains, as they push us to think about the objective functions that are relevant in different behaviours. This helps the construction of general theories that are applicable across domains and highlight both the similarities and differences between the various instantiations of `intelligent' systems.

\section{Experimental Proposal Results}\label{app: exp_proposal_results}

\colorlet{shadecolor}{white}
\definecolor{tab_blue}{HTML}{1f77b4}
\definecolor{tab_orange}{HTML}{ff7f0e}
\definecolor{tab_green}{HTML}{2ca02c}
\definecolor{tab_red}{HTML}{d62728}
\definecolor{summer_green}{HTML}{008066}
\definecolor{boltz_color}{HTML}{e41a1c}
\definecolor{dd_oracle_color}{HTML}{4daf4a}
\definecolor{dd_color}{HTML}{984ea3}
\definecolor{ducb_color}{HTML}{ff7f00}
\definecolor{full_oracle_color}{HTML}{377eb8}

\begin{figure*}[h]      
    \pgfplotsset{
		width=\textwidth,
		height=0.7\textwidth,
		every tick label/.append style={font=\scriptsize},
        y label style={at={(axis description cs:-0.1, 0.5)}},
	}
    \begin{subfigure}[b]{0.48\textwidth}
    \pgfplotsset{
		width=\textwidth,
		height=0.35\textwidth,
		every tick label/.append style={font=\scriptsize},
        y label style={at={(axis description cs:-0.1, 0.5)}},
	}
    \centering
	\begin{tikzpicture}
    \fill[gray, opacity=0] (0, 0) rectangle (\textwidth, 0.65\textwidth);
    \node [anchor=north west] at (0, 0.65\textwidth) {\emph{(a)}};
    \node [anchor=west] at (2.3, 4.7) {\scriptsize No Stimulation};
    \draw[boltz_color, thick] (1.5, 4.7) -- (2.3, 4.7);
    \node [anchor=west] at (5.8, 4.7) {\scriptsize Stimulation};
    \node [anchor=west] at (4.5, 3.8) {\scriptsize + NA Stimulation};
    \draw[full_oracle_color, thick] (5, 4.7) -- (5.8, 4.7);
    \begin{axis}
        [
        xticklabels={},
        xtick={},
        axis line style={draw=none},
        name=main,
        at={(1.3cm,2.5cm)},
        xmin=-100, xmax=4100,
        ymin=-0.05, ymax=1.05,
        ylabel style={align=center},
        ylabel={\small Optimal \\ Arm},
        yticklabels={0, 1},
        ytick={0, 1},
    ]
    \addplot graphics [xmin=-120,xmax=4150,ymin=-0.08,ymax=1.08] {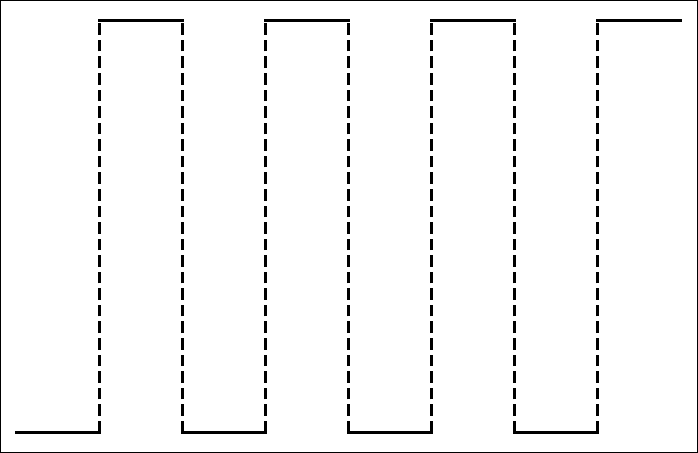};
    \end{axis}
    \begin{axis}
        [
        axis line style={draw=none},
        name=main,
        at={(1.3cm,1cm)},
        xmin=-100, xmax=4100,
        ymin=-0.01, ymax=0.11,
        ylabel style={align=center},
        ylabel={\small Stimulation \\ Protocol},
        xlabel={\small Step},
        yticklabels={0, 0.1},
        ytick={0, 0.1},
    ]
    \addplot graphics [xmin=-120,xmax=4150,ymin=-0.02,ymax=0.12] {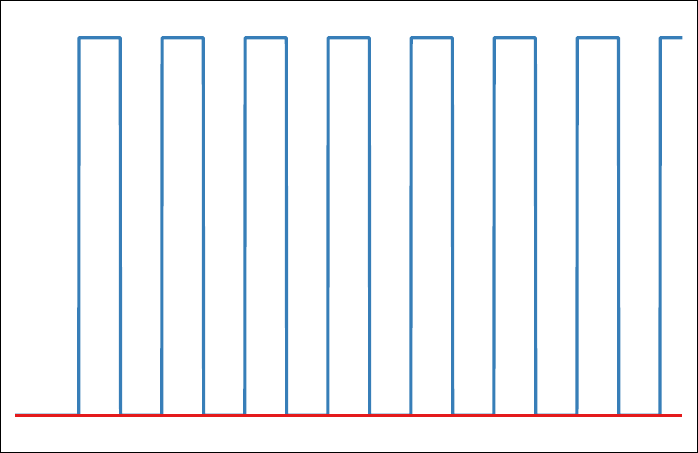};
    \end{axis}
    \end{tikzpicture}%
    \phantomsubcaption
	\label{fig:exp_propa}
	\end{subfigure}
    \begin{subfigure}[b]{0.48\textwidth}
    \centering
	\begin{tikzpicture}
    \fill[gray, opacity=0] (0, 0) rectangle (\textwidth, 0.65\textwidth);
    \node [anchor=north west] at (0, 0.65\textwidth) {\emph{(b)}};
    \begin{axis}
        [
        name=main,
        at={(1.3cm,1cm)},
        xmin=-100, xmax=4100,
        ymin=-0.01, ymax=0.25,
        ylabel={\small Temperature},
        xlabel={\small Step},
    ]
    \addplot graphics [xmin=-100,xmax=4100,ymin=-0.01,ymax=0.25] {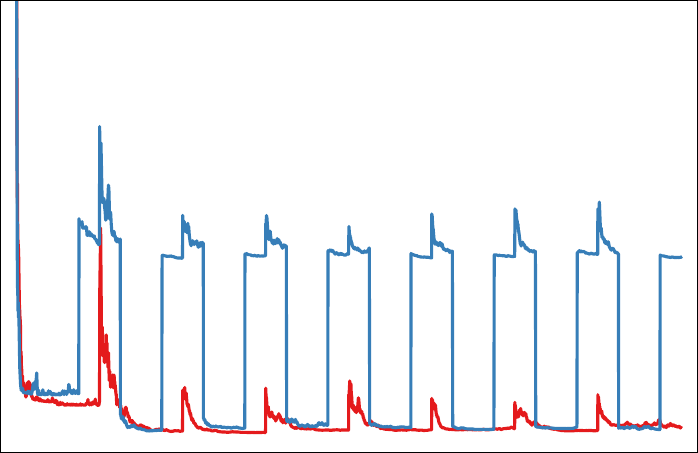};
    \end{axis}
	\end{tikzpicture}%
    \phantomsubcaption
	\label{fig:exp_propab}
	\end{subfigure}
    \begin{subfigure}[b]{0.48\textwidth}
    \centering
	\begin{tikzpicture}
    \fill[gray, opacity=0] (0, 0) rectangle (\textwidth, 0.65\textwidth);
    \node [anchor=north west] at (0, 0.65\textwidth) {\emph{(c)}};
    \begin{axis}
        [
        name=main,
        at={(1.3cm,1cm)},
        xmin=-100, xmax=4100,
        ymin=-0.05, ymax=1.05,
        ylabel={\small Arm Chosen},
        xlabel={\small Step},
    ]
    \addplot graphics [xmin=-100,xmax=4100,ymin=-0.05,ymax=1.05] {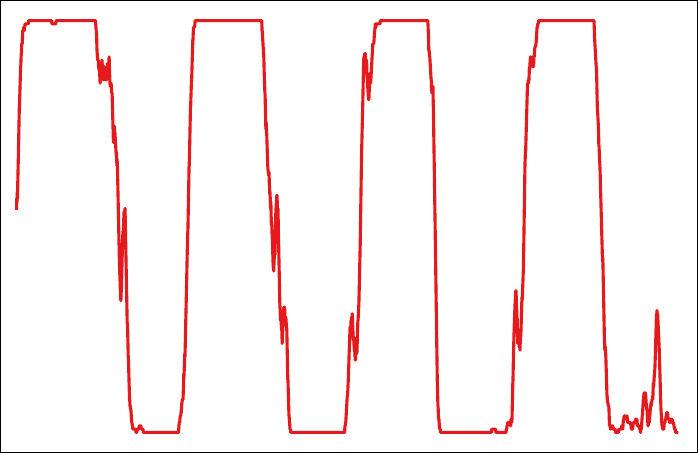};
    \end{axis}
    \end{tikzpicture}%
    \phantomsubcaption
	\label{fig:exp_propc}
	\end{subfigure}
    \begin{subfigure}[b]{0.48\textwidth}
    \centering
	\begin{tikzpicture}
    \fill[gray, opacity=0] (0, 0) rectangle (\textwidth, 0.65\textwidth);
    \node [anchor=north west] at (0, 0.65\textwidth) {\emph{(d)}};
    \begin{axis}
        [
        name=main,
        at={(1.3cm,1cm)},
        xmin=-100, xmax=4100,
        ymin=-0.05, ymax=1.05,
        ylabel={\small Correct Arm},
        xlabel={\small Step },
    ]
    \addplot graphics [xmin=-100,xmax=4100,ymin=-0.05,ymax=1.05] {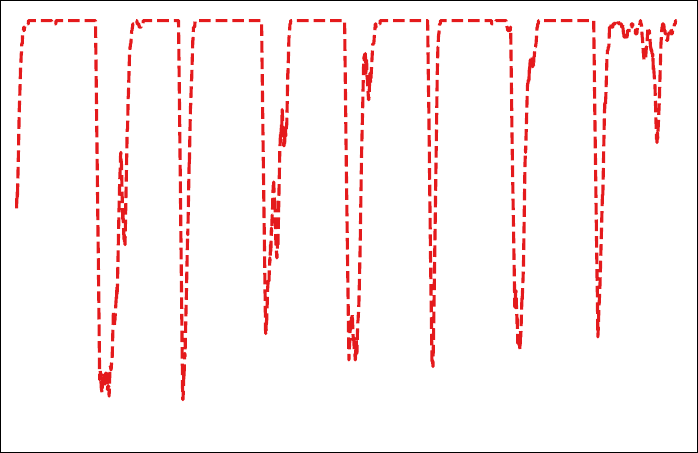};
    \end{axis}
	\end{tikzpicture}%
    \phantomsubcaption
	\label{fig:exp_propd}
	\end{subfigure}
 \begin{subfigure}[b]{0.48\textwidth}
    \centering
	\begin{tikzpicture}
    \fill[gray, opacity=0] (0, 0) rectangle (\textwidth, 0.65\textwidth);
    \node [anchor=north west] at (0, 0.65\textwidth) {\emph{(e)}};
    \begin{axis}
        [
        name=main,
        at={(1.3cm,1cm)},
        xmin=-100, xmax=4100,
        ymin=-0.05, ymax=1.05,
        ylabel={\small Arm Chosen},
        xlabel={\small Step},
    ]
    \addplot graphics [xmin=-100,xmax=4100,ymin=-0.05,ymax=1.05] {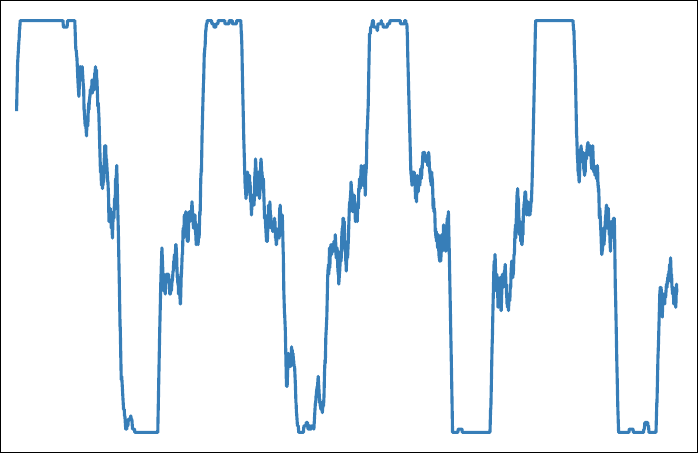};
    \end{axis}
    \end{tikzpicture}%
    \phantomsubcaption
	\label{fig:exp_prope}
	\end{subfigure}
    \begin{subfigure}[b]{0.48\textwidth}
    \centering
	\begin{tikzpicture}
    \fill[gray, opacity=0] (0, 0) rectangle (\textwidth, 0.65\textwidth);
    \node [anchor=north west] at (0, 0.65\textwidth) {\emph{(f)}};
    \begin{axis}
        [
        name=main,
        at={(1.3cm,1cm)},
        xmin=-100, xmax=4100,
        ymin=-0.05, ymax=1.05,
        ylabel={\small Correct Arm},
        xlabel={\small Step },
    ]
    \addplot graphics [xmin=-100,xmax=4100,ymin=-0.05,ymax=1.05] {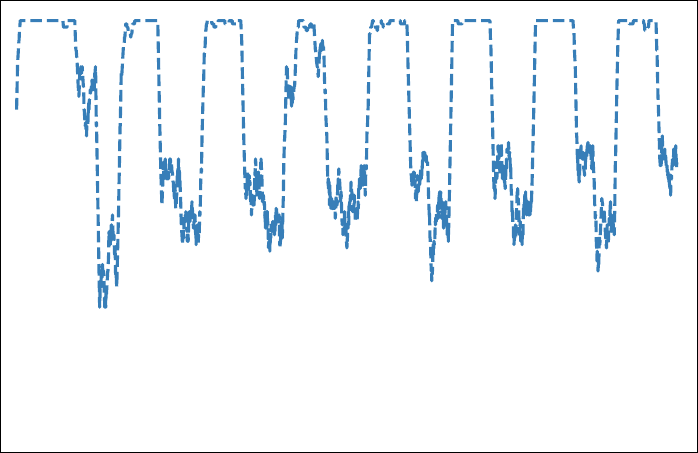};
    \end{axis}
	\end{tikzpicture}%
    \phantomsubcaption
	\label{fig:exp_propf}
	\end{subfigure}
    \caption{\textbf{Experiment Proposal Simulations}. \emph{(a)} simulation schedules for the case of stimulation (blue) and the control with no stimulation (red) \emph{(b)} Temperature trajectory over context switches, stimulation leads to heightened levels of exploration. \emph{(c)/(d)} arm chosen vs. correct arm under no stimulation. \emph{(e)/(f)} arm chosen vs. correct arm under stimulation.}
    \label{fig: app_bandits}
\end{figure*}

Here we present some basic results relating to the experiment proposal presented in~\autoref{sec: proposal}. We compare an oracle Doya-DaYu agent with and without `stimulation' of tonic NA (which under the theories considered has an equivalent computational effect as inhibiting phasic NA) on a two-armed Bernoulli bandit with periodic reversals. By `stimulation' we mean adding a constant to the temperature $\beta^{-1}$.~\autoref{fig:exp_propa} shows the schedule of the `stimulation'~\emph{vis-\`a-vis} the arm probability reversals. In~\autoref{fig:exp_propc}~\&~\autoref{fig:exp_propd} (~\autoref{fig:exp_prope}~\&~\autoref{fig:exp_propf}) we show the action selected by the agent and whether the action is correct for the vanilla (NA Stimulated) Doya-DaYu model. Visually it is clear that the action selection under stimulation is more random; for the plots shown the action breakdown for the stimulated agent is 52\% v.s. 48\% while the non-stimulated agent chooses the `greedy' action 75.9\% of the time in the same period. The temperature of the agents at each step in learning is shown in~\autoref{fig:exp_propab}. Interestingly learning is not always faster despite the higher entropy action selection during a reversal. Clearly during periods of stimulation, the accuracy of the non-stimulated agent is higher: 78\% v.s. 57.6\%, but even measured only over non-stimulated periods the stimulated agent achieves slightly lower accuracy: 98\% v.s. 97\%. One reason for this is that the task is not particularly difficult for the artificial agent and the oracle adaptations are very proficient. Another factor is the tight coupling between exploration and learning rate that results from the expected uncertainty feeding into both quantities. When exploration is higher, overall uncertainty reduces more quickly, which has a downstream effect of lower learning rates. In the control case where the learning rate is kept constant, i.e. not modulated in the way prescribed by Doya-DaYu, we see that the increased entropy does indeed speed up learning. The implication for our experimental predictions is that we would expect to see a higher entropy of action selection under stimulation and a return to `normal' levels outside of stimulation. Meanwhile we would expect the animal to learn the reversal, although it would be difficult (from these preliminary simulations) to make predictions on the speed of that learning since it would require disentangling potential effects of phasic inhibition on the learning rate.
\section{Distributional Reinforcement Learning}\label{app: dist_rl}

While we focus in the main text on a simple bandit task to demonstrate the effectiveness of our neuromodulatory framework, it is also possible to use the same ideas in richer environments. Below we outline ideas for extending the framework to deep learning.

The full~\gls{rl} problem includes high dimensional state and action spaces, and requires more sophisticated methods. Uncertainty estimation has received increasing attention in the context of deep~\gls{rl} in recent years, both in terms of learning general quantities related to return variance~\cite{sherstan2018comparing, dabney2020distributional}, and specifically related to disentangling epistemic and aleatoric uncertainty~\cite{clements2019estimating, charpentier2022disentangling}. Various properties of~\gls{rl} such as non-stationarity and bootstrapping make it a generally difficult problem, but progress has been made. Here, we follow~\citet{clements2019estimating} and ~\citet{jiang2022uncertainty} in using distributional reinforcement learning to perform the relevant uncertainty estimation for the Doya-DaYu agent. 

Distributional reinforcement learning differs from standard approaches to~\gls{rl} in that it attempts to learn the full distribution of returns ($Z^\pi=\sum_{t=0}^\infty \gamma^t R_t$), rather than just the expectation. This makes it a natural candidate for modelling uncertainty in the context of deep reinforcement learning. More concretely, in standard~\gls{rl} the aim is to learn the value function: $Q^\pi(x,a) = \mathbb{E}[Z^\pi(x,a)]$, which is just the first moment of the distribution. A common set of methods used to learn this function involves temporal differences via the so-called Bellman operator, $T$:
\begin{equation}
    T^\pi Q(x,a) := \mathbb{E}[R(x,a)] + \gamma\mathbb{E}_{P,\pi}[Q(x',a')],
\end{equation}
where $P$ governs the state transition probabilities.
There is an analogous operator that acts on the entire distribution, known as the distributional Bellman operator:
\begin{equation}
    T^\pi Z(x,a) :\stackrel{D}{=} R(x,a) + \gamma Z(x',a'), \quad x'\sim P(\cdot|x,a), \quad a'\sim\pi(\cdot|x');
\end{equation}
where $:\stackrel{D}{=}$ signifies that the distributions on the left and right hand sides of the equality are equally distributed. While it is possible to represent $Z^\pi$ discretely using particles spaced along the distribution domain, this induces the necessity of an additional projection of the target onto the original support since $T^\pi Z$ and $Z$ often end up with disjoint supports~\citep{dabney2020distributional}. Thereafter a step is made to minimise the KL divergence to the projected target. Not only does this unwieldy projection step introduce further approximation, the distributional Bellman operator is~\emph{not} a contraction in KL divergence.

A more principled approach (that is also empirically superior) is to represent the distribution it in terms of~\emph{quantiles}~\citep{dabney2018distributional} such that everything is defined in the space of the inverse cumulative distribution of the return. From there we can employ a method known as~\emph{quantile regression loss}, which directly minimises a Wasserstein loss with unbiased gradients (which is not possible under the particle formulation). Among other empirical advantages, the distributional Bellman operator is a contraction in the Wasserstein space so we are guaranteed to move towards a fixed point in the landscape!

The above basics allow us to present methods for estimating epistemic and aleatoric uncertainty. More comprehensive treatment of the subject of distributional~\gls{rl} can be found in~\citet{bellemare2023distributional}.

\citet{clements2019estimating} define epistemic and aleatoric uncertainty in interactive environments as:
\begin{align}
    \sigma_{\text{epistemic}}^2 &= \mathbb{E}_{i\sim\mathcal{U}\{1, N\}}[\text{var}_{\mathbf{\theta}\sim P(\mathbf{\theta}|D)}(y_i(\mathbf{\theta}; s, a))],\\
    \sigma_{\text{aleatoric}}^2 &= \text{var}_{i\sim\mathcal{U}\{1, N\}}[\mathbb{E}_{\mathbf{\theta}\sim P(\mathbf{\theta}|D)}(y_i(\mathbf{\theta}; s, a))];
\end{align}
where $i$ is the quantile index (of the estimated return distribution), $\mathcal{U}$ is the uniform distribution, $\theta$ are the model parameters, and $y$ is the quantity of interest e.g. a return. We get access to the distribution of model parameters at a given quantile, $P(\theta|D)$, by using an ensemble of distributional agents. 
An example instance of an algorithm in this direction can be found in~\citet{jiang2022uncertainty}.

\subsection{Key-Door Gridworld Environment}\label{app: key-door}

\begin{figure*}[t]
    \begin{subfigure}[b]{\textwidth}
    \includegraphics[width=\textwidth]{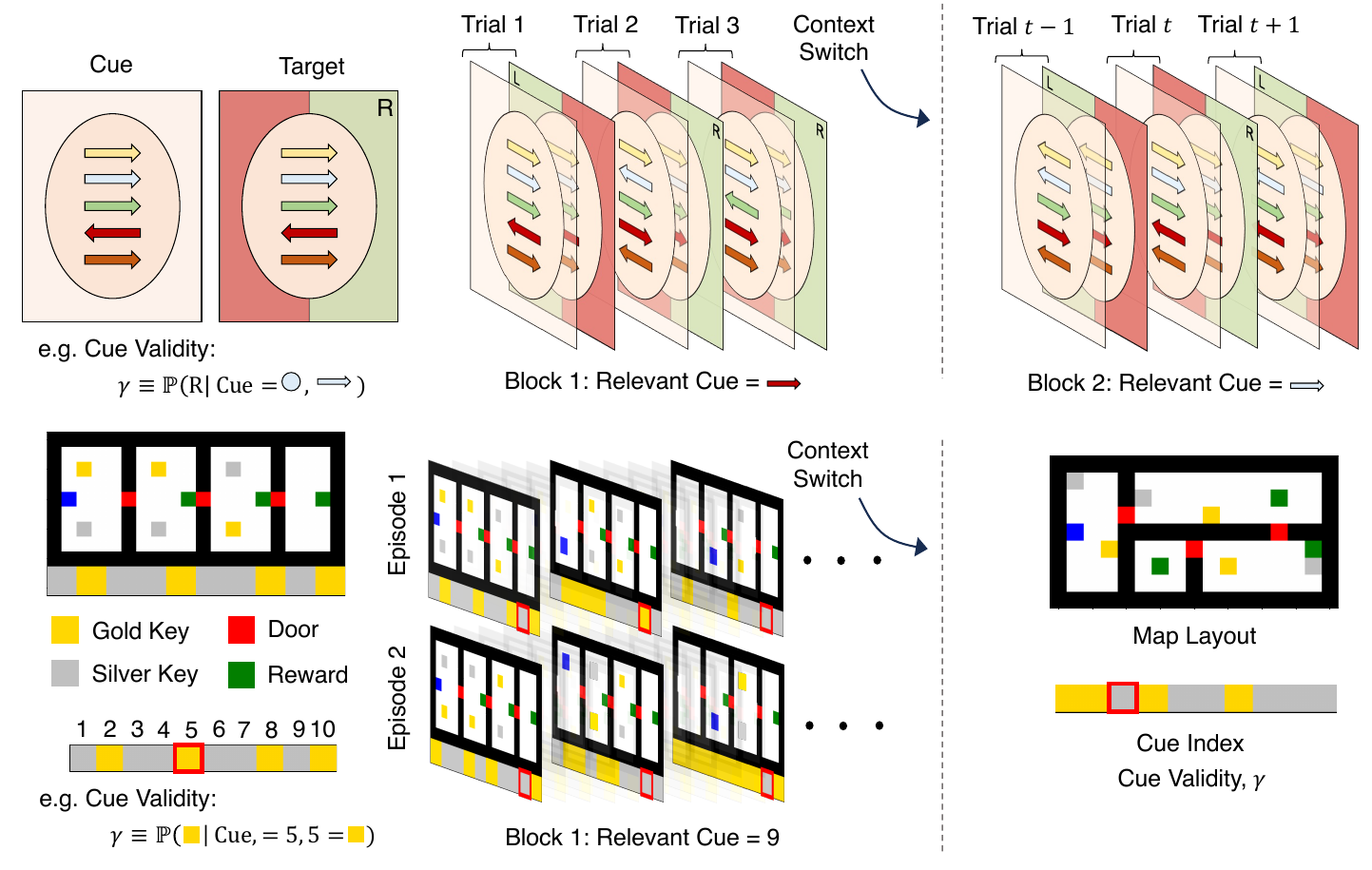}
    \centering
    \phantomsubcaption
	\label{fig:1a}
	\end{subfigure}
    \begin{subfigure}[b]{\textwidth}
    \centering
    \phantomsubcaption
	\label{fig:1b}
	\end{subfigure}
    \caption{\textbf{Posner Key-Door Environment}. \emph{(a)} In the \emph{extended Posner task}~\cite{dayan2002expected}, participants must model two types of uncertainty: the validity of the cue (`expected uncertainty'), which encodes how likely the cue arrow on the stimulus points in the target direction; and the possibility of a context switch in which the cue index and validity may change (`unexpected uncertainty'). \emph{(b)} In our navigation~\emph{key-door} environment, agents must collect the correct key in each room to open doors and collect rewards in ensuing rooms. In each room a cue indicates the correct key to collect with some validity. At a context switch, the map layout (wall, key, door and reward locations), cue index, cue validity, and block length may change.}
    \label{posner}
\end{figure*}

To test our framework and Doya-DaYu agents in a function approximation setting, we introduce here a
novel~\gls{rl}~environment, based around navigation. It is heavily inspired by the
so-called extended Posner task modelled by~\citet{dayan2002expected}. In each trial of the extended Posner task (illustrated in~\autoref{fig:1a}), participants are shown a stimulus from which they must make a binary left-right decision; this is followed by a target, which reveals the correct direction. The stimulus consists of a set of arrows pointing either left or right. One of the arrows is the~\emph{cue}, which with a certain probability $\gamma$ (known as the~\emph{cue validity}) points in the direction of the target. All other arrows are distractors and point in random directions. Multiple trials make up a block, during which the cue (arrow indicating target direction) and the cue validity remain the same. At some unknown time, there is a context switch, leading to a new block of trials with potentially a different cue index and cue validity. Thus two sources of randomness need to be modelled to effectively solve the task: the probabilistic nature of the cue, and the possibility of a context switch. 

Our~\emph{key-door} environment (illustrated in~\autoref{fig:1b}) consists of multiple rooms, separated by doors. All but the final room contain two keys: one silver and one gold. All but the first room contain a reward. If the agent picks up either key in a given room, the door to the next room opens and the other key vanishes. However, if the `wrong' key is collected, the reward in the next room also vanishes. Additionally---in analogy to the Posner task---there is a row of pixels beneath the maze, one of which is the cue that with a given probability (cue validity) indicates the color of key that should be collected in the next room. Again, periodically at random, there is a context switch in which changes can occur in the cue index, cue validity, and now too the map layout.

\section{Supplementary Experimental Methods}\label{app:exp_proposal_full}

Here we provide additional details of the experimental set-up for the study proposed in~\autoref{sec: proposal}. 

\subsection{Task set-up}

To enable high-quality neural recordings from multiple trials with high engagement, mice are head-fixed and water-deprived. Care is taken to prevent their weight from dropping below 80\% of its value before the experiment. Arm rewards follow fixed reward magnitude (e.g., 3$\mu l$) Bernoulli distributions, with success probabilities summing to 1 across arms, thereby avoiding assumptions about mice's perceived value of different water amounts---with comparisons only between probabilities of reward presence or absence across arms. 

\subsection{Trial timing}

At the start of each trial, an auditory `go' cue is presented to indicate that a decision can be made (\autoref{fig:4c}). The mouse then uses its tongue to choose a port, which detects the lick using a capacitive-sensing circuit board connected to the lick ports~\citep{duan2021cortico}. Upon detection, the port delivers a water reward sampled from its corresponding distribution; the port then retracts to avoid the animal making more than one lick and thus enabling adequate credit assignment. We record both the side of the chosen port as well as the time taken to make that choice (i.e. time from `go' cue to lick detection).

\subsection{NA manipulation}

We note that there are other methods of manipulating NA levels (e.g., pharmacological or chemogenetic) which could also be used to tackle the question here described. However, we believe that optogenetic manipulation is the most suitable due to its reversible nature and cell-type specificity.

\end{document}